%% file: main.tex
\documentclass[letterpaper, 10 pt, conference]{ieeeconf}  %
\IEEEoverridecommandlockouts                              %
\usepackage{fp}
\FPset{\pb}{0}
\newcommand{\pagebudget}[1]{}
\newcommand{\showtotalpagebudget}[1]{}

\renewcommand*{\gobble}[1]{}

\usepackage{soul} %
\setstcolor{blue}

\usepackage[dvipsnames]{xcolor}
\usepackage{pgf,tikz,pgfplots}
\PassOptionsToPackage{dvipsnames,table}{xcolor}
\pgfplotsset{compat=1.15}
\usepackage{mathrsfs}
\usetikzlibrary{arrows,backgrounds}
\usepackage[style=ieee,backend=biber,dashed=false,doi=false,eprint=false, sorting=nyt]{biblatex}
\bibliography{main}
\usepackage{pdfpages}
\usepackage{algorithmicx}
\usepackage{algorithm}
\usepackage[noend]{algpseudocode}
\usepackage{amsmath} 
\usepackage{amssymb} 
\usepackage{array}
\usepackage{epsfig}
\usepackage{enumerate}
\usepackage{graphicx} 
\usepackage{upgreek}
\usepackage[english]{babel}
\usepackage{csquotes}   %
\usepackage[normalem]{ulem}
\usepackage{xspace}
\usepackage{tabularx}
\usepackage{pseudocode}
\usepackage{subfloat}
\usepackage{svg}
\usepackage{tikz}

\algrenewcommand\algorithmicrequire{\textbf{Precondition:}}
\algrenewcommand\algorithmicensure{\textbf{Postcondition:}}
\algnewcommand\algorithmicinput{\textbf{Input:}}
\algnewcommand\Input{\item[\algorithmicinput]}
\algnewcommand\algorithmicoutput{\textbf{Output:}}
\algnewcommand\Output{\item[\algorithmicoutput]}
\newtheorem{theorem}{Theorem}[section]
\newenvironment{definition}[1][Definition]{\begin{trivlist}
\item[\hskip \labelsep {\bfseries #1}]}{\end{trivlist}}
\newcolumntype{C}{>{\centering\arraybackslash} m{2.2in} }

\renewcommand{\Re}{\mathbb{R}}

\newcommand{\old}[1]{{}}
\makeatletter
\newcommand{\Rmnum}[1]{\expandafter\@slowromancap\romannumeral #1@}
\makeatother

\usepackage{support-caption}   %
\usepackage[font=footnotesize]{caption}%
\makeatletter
\def\thm@space@setup{%
  \thm@preskip=.5pt plus .5pt minus .5pt
  \thm@postskip=\thm@preskip %
}
\makeatother

\title{\LARGE \bf
A Visibility Roadmap Sampling Approach for a Multi-Robot Visibility-Based Pursuit-Evasion Problem
}

\author{Trevor Olsen, Anne M. Tumlin, Nicholas M. Stiffler and Jason M. O'Kane%
\thanks{%
	T. Olsen, A. M. Tumlin, N. M. Stiffler, and J. M. O'Kane are with the Department
    of Computer Science and Engineering, University of South Carolina, Columbia, SC 29208, USA. 
    {\tt \footnotesize \{tvolsen, atumlin\}@email.sc.edu \{stifflen, jokane\}@cse.sc.edu} %
	This material is based upon work supported by the National Science Foundation 
    under Grant Nos. 1659514 and 1849291.%
}}

\begin{document}

\maketitle
\thispagestyle{empty}
\pagestyle{empty}

\begin{abstract}
Given a two-dimensional polygonal space, the multi-robot visibility-based pursuit-evasion problem tasks several pursuer robots with the goal of
establishing visibility with an arbitrarily fast evader. The best known complete algorithm for this problem takes
time doubly exponential in the number of robots. However, sampling-based techniques have shown promise in 
generating feasible solutions in these scenarios. 
One of the primary drawbacks to employing existing sampling-based methods is that existing algorithms have long execution
times and high failure rates for complex environments. This paper addresses that limitation by proposing a new algorithm that takes an
environment as its input and returns a
joint motion strategy which ensures that the evader is captured by one of the pursuers. Starting with a single pursuer, we sequentially construct
Sample-Generated Pursuit-Evasion Graphs to create such a joint motion strategy. This sequential graph structure ensures that our
algorithm will always terminate with a solution, regardless of the complexity of the environment. We describe an implementation of this algorithm and
present quantitative results that show significant improvement in comparison to the existing algorithm.  
\end{abstract}

\section{Introduction}\pagebudget{1}

Autonomous reconnaissance tasks, in which robots strive to observe salient features of objects or agents within their environments, remain
one of the most active threads of research within the robotics community.
Such tasks have wide-ranging application domains such as
environmental monitoring~\cite{DuaGom+16,DunMar12,IslNoo+15,TiwCho19,TokBha+10}, surveillance~\cite{AceBeg+14,AlaRah+17,DamTokKum17,FenHan+19},
and search-and-rescue~\cite{HolKehSin07,StiOKa11,MimBerMar20}.
Many of these tasks can be framed as two-player games played amongst opposing teams: evaders (who wish to evade capture)
and pursuers (who seek to capture them).
This paper is concerned with a  specific form of this two-player game, wherein a team of pursuers must locate an evader (or group of evaders) in a polygonal environment. %

\begin{figure}[t]
    \centering
    \includegraphics[width=.8\columnwidth, keepaspectratio]{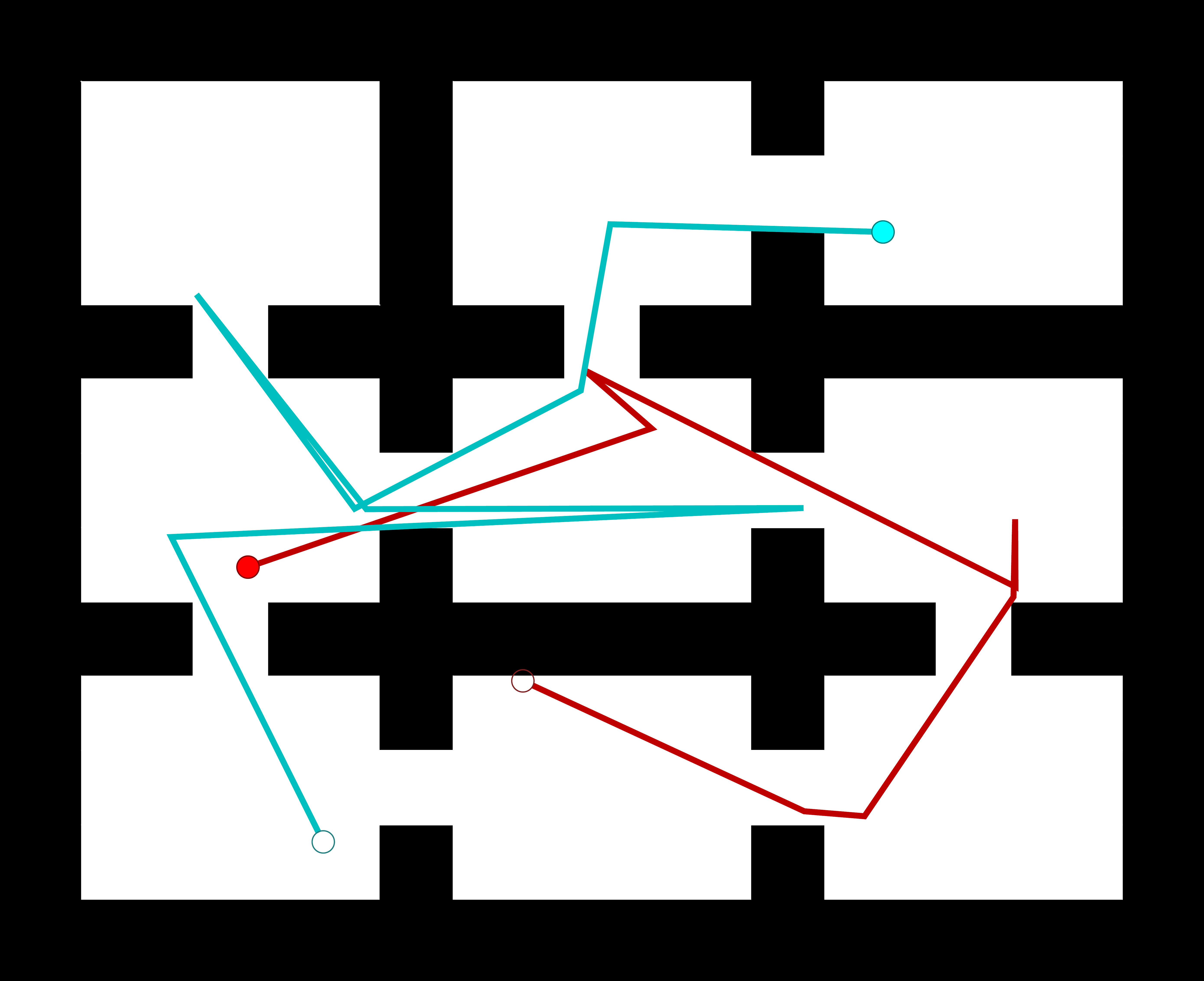}
    \caption{An example multi-pursuer solution strategy generated by the proposed algorithm. The filled circle represent the initial location of the pursuer, while the hollow circles represent its final location.}
    \label{fig:first_page}
\end{figure}

Specifically, we address one such problem where a group of pursuers, each equipped with an omni-directional sensor that extends to the polygonal boundary, must form a motion plan to locate an arbitrarily fast evader in a polygonal environment. Figure~\ref{fig:first_page} illustrates this scenario.
The literature has a number of results for this problem in the single-pursuer case, including algorithms with strong guarantees such as
completeness~\cite{GuiLat+99} and path length optimality~\cite{StiOKa17}. %
However, the case in which multiple pursuers cooperate is not nearly as well understood.  A complete algorithm is known, but it runs in time doubly-exponential in the number of pursuers~\cite{StiOKa14a}.

One approach to overcome the computational challenge posed by this multiple-pursuer pursuit-evasion planning task%
~\cite{StiOKa14b}, showed the feasibility of sampling-based techniques to generate joint motion strategies.  Nevertheless, that approach has several important limitations.
\begin{enumerate}
    \item[\footnotesize{(i)}]
    The existing algorithm lacks insight into how the sampling should be performed, and treats the sampling distribution as a ``black box.''%
    
    \item[\footnotesize{(ii)}]
    The existing algorithm requires a predetermined number of pursuers as input; it cannot adapt the number of pursuers to the complexity of the environment. 
    
    \item[\footnotesize{(iii)}]
    The solutions generated by this approach are of poor quality, in the sense that there nearly always are motions by one or more of the pursuers that do not actively contribute to the search. 
\end{enumerate}
After a review of related work (Section~\ref{sec:related}), precise statement of the problem (Section~\ref{sec:ps}), and a summary of important concepts from prior work on this problem (Section~\ref{sec:background}), this paper makes three new contributions that address the limitations of the existing algorithm.
\begin{enumerate}
    \item[\footnotesize{(i)}]
    We introduce a new sampling strategy, tailored to the visibility-based nature of the problem (Section~\ref{sec:web}).

    \item[\footnotesize{(ii)}]
    We describe a method that eliminates the need for the number of pursuers to be provided as input, instead iteratively increasing the size of the team 
    (Section~\ref{sec:recursive}).
    
    \item[\footnotesize{(iii)}]
    We present a post-processing algorithm that improves solution quality
    (Section~\ref{sec:improve}).
\end{enumerate}
Finally, Section~\ref{sec:experiments} presents quantitative evaluations of these new improvements and Section~\ref{sec:conclusion} previews future work.

\section{Related work}\label{sec:related}
\pagebudget{0.4}
This research blends ideas from two vibrant threads of prior robotics research: visibility-based pursuit-evasion and sampling-based motion planning in high dimensional spaces.

In regard to pursuit-evasion, this work is most closely aligned with the %
problem first introduced by Suzuki and Yamashita~\cite{SuzYam92}, in which an evader
operating in a geometric environment seeks to locate an unpredictable evader
capable of moving arbitrarily quickly.
This work was later expanded by Guibas, Latombe, LaValle, Lin, and
Motwani~\cite{GuiLat+99} who provide a complete algorithm for the single pursuer
scenario in simply-connected environments for a pursuer with an omnidirectional
field-of-view. Park, Lee, and Chwa~\cite{ParLeeChwa01} identified
necessary and sufficient conditions for a search to be feasible for a single pursuer.

Other results for the single pursuer scenario that build upon this foundation provide results such as completeness~\cite{GuiLat+99}, optimality~\cite{StiOKa17}, establishing and maintaining visibility of a moving agent~\cite{SkhDud17, SkhKakDud18}
or consider more restrictive scenarios with respect to pursuer parameters such as sensing, actuation and speed~\cite{SacLavRaj04, GerThrGor06, TovLav08,RajLav01,StiOKa16, SkhDud13}.

The community has placed increased emphasis on the study of
richer scenarios where a team of pursuers cooperate during the search~\cite{KolCar10a, DurFraBul12,GreGiv+17, StiOKa14b, IslKanKha05, KolCar07, GanCorBul07}.
Stiffler and O'Kane~\cite{StiOKa14a} present an algorithm utilizing a cylindrical algebraic decomposition that, while complete, relies upon constructing a graph whose size is doubly exponential in the complexity of the environment.
That work subsequently served as motivation for approaches that utilize heuristics~\cite{StiOKa14b} that seek to overcome the problem complexity by utilizing sampling techniques.

More generally, sampling based techniques have been employed in a number of planning contexts where computing an exact solution proves computationally intractable such as 
motion planning~\cite{KavSve+96, MarBek13, LavKuf00,SolSalHal14,KarFra11} and 
manipulation planning~\cite{SimLau+04,Hau14,KroBek16}.
One caveat of sampling-based methods is that they quite often suffer from the \textit{curse of dimensionality}~\cite{Bel57} whereby, as the number of dimensions increases, the search space becomes so vast that the number of samples required for adequate coverage of the space increases dramatically.
A number of different approaches have been proposed to combat this problem.
One recent result draws samples in lower-dimensional subspaces to search for a feasible solution, and
incrementally reasons about higher dimensions while utilizing the information gained in the lower dimensional graph~\cite{XanEsp+20}.
For the specific multi-robot case in which the configuration space is a Cartesian product of the configuration spaces of individual agents in the system, one novel approach seeks to reason about each agent independently
(a subdimension), and only when the agents reach a point where they interact with one
another is there a lifting to a higher-dimensional space~\cite{WagKanCho12}. 

\section{Problem statement}\label{sec:ps}\pagebudget{0.7}

\subsection{The environment, the evader, and the pursuers}
Let $E \subset \mathbb{R}^2$ be a bounded, closed, connected polygonal set called the \emph{environment}.
An evader moves within $E$.
We describe the evader's position via a continuous function $e(t): [0,\infty) \rightarrow E$ to represent the location of the evader at time $t$.
There is no restriction on the speed of the evader, so long as it is finite. 

We task $n$ pursuers with the goal of ensuring capture of the evaders regardless of the evader trajectories. We can assume worst case reasoning in the sense that there is a singular evader which follows a path that maximizes its time until detection over all evader trajectories.
We consider both scenarios in which the number of pursuers is known and fixed, an in which the number of pursuers to utilize is determined by the algorithm as the plan is generated.
Let $p_i: [0,\infty) \rightarrow E$ represent the location of the $i^\text{th}$ pursuer as a function of time. We refer to such functions as \emph{motion strategies}. Each pursuer is equipped with knowledge of $E$ and an omnidirectional sensor which extends until the nearest point on the boundary of $E$ in each direction. That is, a pursuer at point $q_1 \in E$ can see every point in its visibility polygon, $V(q_1) = \{q_2 \in E \mid  \overline{q_1 q_2} \subseteq E\}$ where $\overline{q_1 q_2}$ denotes the line segment joining $q_1$ and $q_2$ in $E$.

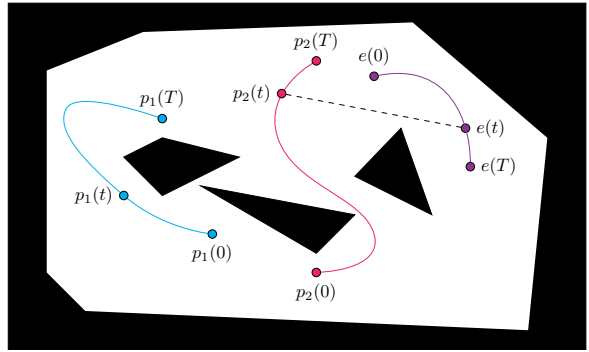
\begin{figure}[t]
\centering
\resizebox{0.9\columnwidth}{!}{
\begin{tikzpicture}[scale=0.75, yscale=0.5]

\fill[draw=black,fill=black] (-9,-5) rectangle (6,13);

\fill[color=white, line width=0.3mm] (-5.5,11.5)--(-8,9.5)--(-8,-1)--(-7,-3)--(4.5,-4)--(5,6)--(1.5,12)--(-5.5,11.5);

\draw [cyan] plot [smooth, tension=1] coordinates {(-3.7,1) (-6, 3) (-7.5, 7.5) (-5,7)};
\draw [WildStrawberry] plot [smooth, tension=1] coordinates {(-1,-1) (0.5,1) (-2,6) (-1,10)};

\fill[draw=black,fill=black] (-5,6) -- (-6,5) -- (-5,3) -- (-3,5) -- cycle;
\fill[draw=black,fill=black] (1.2,6.5) -- (0,4) -- (2,2) -- cycle;
\fill[draw=black,fill=black] (-1,0) -- (-4,3.5) -- (0,2) -- cycle;
\node[fill=cyan, circle, draw=black, scale=0.5, label=below:{$p_1(0)$}] (p1) at (-3.7,1) {};
\node[fill=cyan, circle, draw=black, scale=0.5, label=left:{$p_1(t)$}] (pt) at (-6, 3) {};
\node[fill=cyan, circle, draw=black, scale=0.5, label=above:{$p_1(T)$}] (p2) at (-5,7) {};
\node[fill=WildStrawberry, circle, draw=black, scale=0.5, label=below:{$p_2(0)$}] (q1) at (-1,-1) {};
\node[fill=WildStrawberry, circle, draw=black, scale=0.5, label=left:{$p_2(t)$}] (qt) at (-1.9,8.3) {};
\node[fill=WildStrawberry, circle, draw=black, scale=0.5, label=above:{$p_2(T)$}] (q2) at (-1,10) {};
\node[fill=Fuchsia, circle, draw=black, scale=0.5, label=above:{$e(0)$}] (e1) at (0.5,9.2) {};
\node[fill=Fuchsia, circle, draw=black, scale=0.5, label=right:{$e(t)$}] (et) at (2.875,6.5) {};
\node[fill=Fuchsia, circle, draw=black, scale=0.5, label=right:{$e(T)$}] (e2) at (3,4.5) {};

\draw [color=black,dashed] (qt)--(et);

\draw [color=Fuchsia] (e1) to [out=20,in=90] (e2);

\end{tikzpicture}}
\caption{Example evader and pursuer paths.  The evader is detected at time $t$.}  %
\label{fig:notation}
\end{figure}

\subsection{Objective}
The algorithmic problem we address is to find a collection of motion strategies  $\{ p_1, p_2, \dots , p_n \}$ such that, for some $t \geq 0$ and $j \in \{1, \dots , n\}$, $e(t) \in  V(p_j(t))$ for any evader curve $e$. Such a collection of motion strategies is called a \emph{solution}.
Specifically, we consider two related problems.

\paragraph{Fixed} Given an environment $E$ and a positive integer $n$, generate a solution using exactly $n$ pursuers.  Algorithms for this problem can be evaluated by examining both the time needed to generate a solution as well as the time needed for the pursuers to execute that solution.
    
\paragraph{Variable} Given an environment $E$, generate a solution that uses as few pursuers as possible.  In addition to the run time and execution time criteria mentioned above, algorithms for this variant of the problem can also be judged by the number of pursuers utilized by the computed solution.

\section{Background}\label{sec:background}
\pagebudget{0.9}
This section concisely summarizes some essential prior results upon which our new contributions build.  Specifically, we describe 
how the pursuers' knowledge about the evader's possible position can be maintained (Section~\ref{ssec:shadows}) and present a data structure that encapsulates the progress of a search for a complete solution (Section~\ref{ssec:sg-peg}).

\subsection{Shadows and shadow events}\label{ssec:shadows}

In general, the pursuers will be able to see only a subset of the environment at any particular time, while the remaining parts of the environment are out of view of all of the pursuers. Formally, at time $t$, the \emph{shadow region}, $S(t)$, is $E \setminus \bigcup_{i=1}^{n} V(p_i(t))$. We call the maximally path-connected components of $S(t)$ \emph{shadows}.

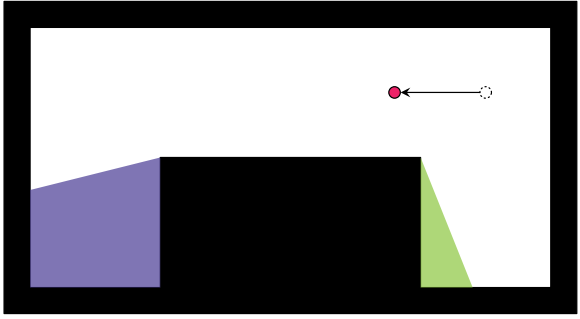
\begin{figure}[t]
\centering
\resizebox{0.9\columnwidth}{!}{
\begin{tikzpicture}[line cap=round,line join=round,>=triangle 45,x=1cm,y=1cm, scale=1.5]
\centering
\fill[draw=black,fill=black] (-0.2,1.2) rectangle (4.2,-1.2);
\fill[draw=black,fill=white] (0,1) -- (4,1) -- (4,-1) -- (3,-1) -- (3,0) -- (1,0) -- (1,-1) -- (0,-1) -- (0,0) -- cycle;
\fill[fill=LimeGreen, opacity=0.7] (3,0) -- (3.4,-1) -- (3,-1) -- cycle;
\fill[fill=BlueViolet, opacity=0.7] (0,-0.25) -- (0,-1) -- (1,-1) -- (1,0) -- cycle;

\node[draw=black, fill=WildStrawberry, circle, scale=0.4] (a) at (2.8,0.5) {};
\node[draw=black, style=densely dotted, circle, scale=0.4] (b) at (3.5,0.5) {};
\draw [->, >=stealth] (b)--(a);
\end{tikzpicture}}
\caption{An example of a cleared (right) and contaminated (left) shadow, given the pursuer's movement history. Here, the white region displays the visibility polygon of a pursuer located at the red circle.}
\end{figure}

The crucial information about each shadow is whether or not the evader may be hiding within that shadow. A shadow $s$ is 
called \textit{cleared} at time $t$ if, based on the pursuers' joint motions up to time $t$, it is not possible for the evader to be within 
$s$ without having been seen by one of the pursuers. Analogously, a shadow is said to be \textit{contaminated} if it is possible for 
the evader to be hiding within it given the pursuers' motions up to time $t$.
As the pursuers move through the environment, the individual shadows change continuously. However, the cardinality of the set of shadows
changes only when a \emph{shadow event} occurs, i.e. a shadow \textit{appears}, \textit{disappears}, \textit{splits}, or \textit{merges} with
another shadow. 
The shadow events induce the following changes to the
status of a shadow:
\begin{itemize}
    \item \emph{Appear}: A shadow can appear if the pursuers lose vision of a region within the environment. In this event, the new shadow has a cleared status.
    \item \emph{Disappear}: A shadow can disappear if the pursuers gain vision of a region which was previously a shadow. The label corresponding to the shadow that disappears is discarded.
    \item \emph{Merge}: Two or more shadows can merge into a larger shadow. In this event, the new shadow is assigned the label cleared only if all merging shadows were cleared; otherwise, the new shadow is labeled contaminated.
    \item \emph{Split}: If a shadow becomes path disconnected, we say it was split. The newly formed shadows have the status of the initial from which they were formed.%
\end{itemize}
More detail about these types of shadow events and their clear/contaminated labels may be found in the work of Guibas, Latombe, LaValle, Lin, and Motwani~\cite{GuiLat+99}.

\subsection{Sample-generated pursuit-evasion graphs}\label{ssec:sg-peg}

To aid in the search for a joint motion strategy for the pursuers, we utilize the Sample-Generated Pursuit-Evasion Graph (SG-PEG) data structure; additional details about the SG-PEG appear in the original paper~\cite{StiOKa14b}.
An SG-PEG is a directed graph $G=(V_G, E_G)$, representing a portion of the connectivity of the joint configuration space for a fixed number $n$ of pursuers.
Each vertex in $V_G$ represents a joint configuration for the pursuers 
    $(p_1, \dots, p_n) \in E^n$.
Edges in $E_G$ connect pairs of vertices for which it is possible for each pursuer to make a collision-free straight line motion between the two corresponding configurations.
That is, if an edge exists between two vertices representing joint configurations $(p_1, \dots , p_n)$ and $(q_1, \dots, q_n)$, then $\overline{p_iq_i} \subset E$ for each $i \in \{1,\ldots,n\}$.
One vertex, corresponding to the starting joint configuration of the robots, is designated as the \emph{root} of the graph.  Each vertex maintains a list of non-dominated shadow labels (i.e. clear/contaminated statuses) that are reachable by traversing the graph along some (possibly non-simple) path from the root.

The primary operation that can be performed on a standard SG-PEG is \textsc{AddSample}$(p_1,\dots,p_n)$ which, given a collision-free joint configuration, performs three main steps:
\begin{itemize}
    \item[\footnotesize{(i)}] It inserts a new vertex $v$ at the given joint configuration.

    \item[\footnotesize{(ii)}] For any other vertex $u$ within a pre-defined connection distance in $E^n$, it checks whether the straight-line connection between $v$ and $u$ would be collision-free.  If so, and if the shortest path in the graph between $v$ and $u$ is sufficiently long, it creates edges $vu$ and $uv$.
    
    \item[\footnotesize{(iii)}] For each new edge thusly added, \textsc{AddSample} computes the shadow events described in Section~\ref{ssec:shadows} induced by motion along that edge.  It then propagates the reachable shadow label information across the new edge and then recursively across the graph, to determine what new reachable shadow labels, if any, are now possible at which vertices, due to this new edge.
\end{itemize}
The utility of the SG-PEG is that if any vertex $v$ has a reachable shadow label that is fully cleared (i.e. it has a reachable shadow label set in which each shadow bears a clear label), then a solution can be extracted from the graph by tracing back via the appropriate edges to the root vertex.

\section{Algorithm description}\label{sec:ad}
Our approach to this problem is based on two significant additions and one modification to the prior algorithm of Stiffler and O'Kane \cite{StiOKa14b}. 
The basic idea of that prior algorithm is to generate random samples in $E^n$ and use them to construct an SG-PEG, continuing until the SG-PEG indicates that it contains a solution.
Algorithm~\ref{algo:solve} shows the enhancements that we propose.  New elements in comparison to the prior algorithm are highlighted: modifications to the sampling strategy in purple text and new additions in blue text. Note that, $n$, the number of robots has been \sout{removed} in our variant.
Details about these changes appear below.

\begin{algorithm}[t]
  \small
  \caption{\textsc{Solve}($E$, \sout{$n$}, \textcolor{blue}{$C$}, \textcolor{purple}{$S$})}
  \begin{algorithmic}[1]
    \Input{an environment $E$, \sout{a number of pursuers $n$},\newline{}\hspace*{4mm} \textcolor{blue}{an expansion criterion $C$} and \textcolor{purple}{a sampler $S$}}
    \State $G \gets $ empty SG-PEG for \sout{$n$ pursuers} \textcolor{blue}{1 pursuer}
    \State $p \gets \textcolor{purple}{\textsc{S.getSample}()}$\label{line:sample1}
    \State $G.\textsc{addSample}(p)$
    \State $G.\textsc{setRoot}(p)$
    \While{no solution has been found}
        \State $p \gets \textcolor{purple}{S.\textsc{getSample}()}$\label{line:sample2}
        \State $G.\textsc{addSample}(p)$
        \If{\textcolor{blue}{$C$ is met}}\label{line:term}
            \State $\textcolor{blue}{G.\textsc{addPursuer}()}$\label{line:clone} 
        \EndIf
    \EndWhile
    \State $X \gets \textsc{extractSolution}()$
    \State $\textcolor{blue}{X' \gets \textsc{refineSolution}(X)}$
    \State \Return $X'$

  \end{algorithmic}
  \label{algo:solve}
\end{algorithm}

\subsection{Web sampling}\label{sec:web}
\pagebudget{0.75}
Stiffler and O'Kane proposed a handful of sampling distributions, but none of them proved significantly more effective than simple uniform random sampling of the joint configuration space.
This approach appears not to be particularly effective in this domain, because the environment is comprised of regions whose
surveyance is essential to finding a solution.
To combat this issue, we propose a new strategy for generating samples, which specifically takes into account the visibility component of the problem. 
Specifically, the goal is to generate a small collection of samples $W$ that has two properties: (a) each point in $E$ is seen by at least one point in $W$, and (b) a pursuer moving between points in $W$ along straight line segments in $E$ can travel between any pair of points in $W$, i.e. $W$ forms a `connected roadmap' of $E$.

\begin{figure}[t]

    \includegraphics[width=0.47\columnwidth]{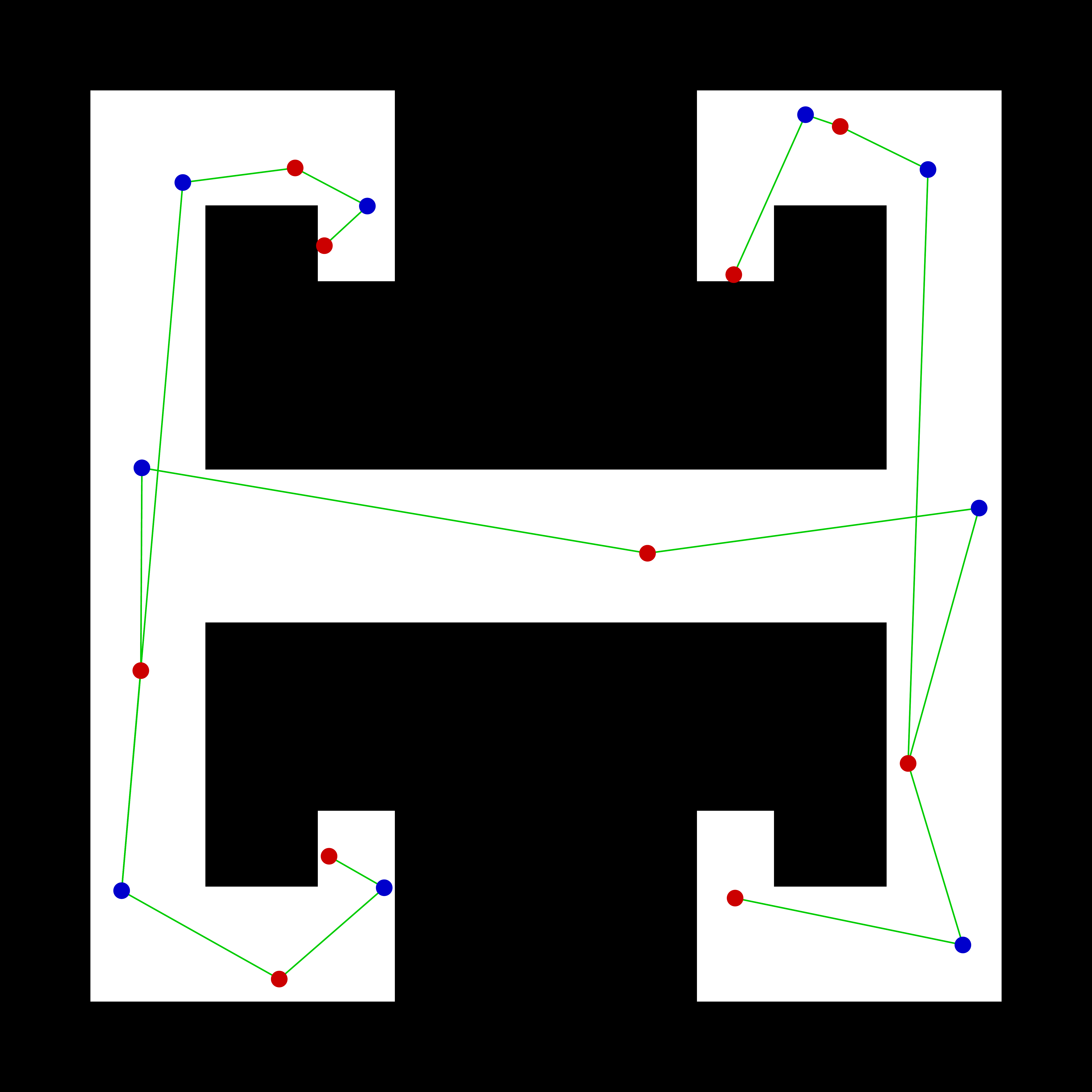}
    \includegraphics[width=0.47\columnwidth]{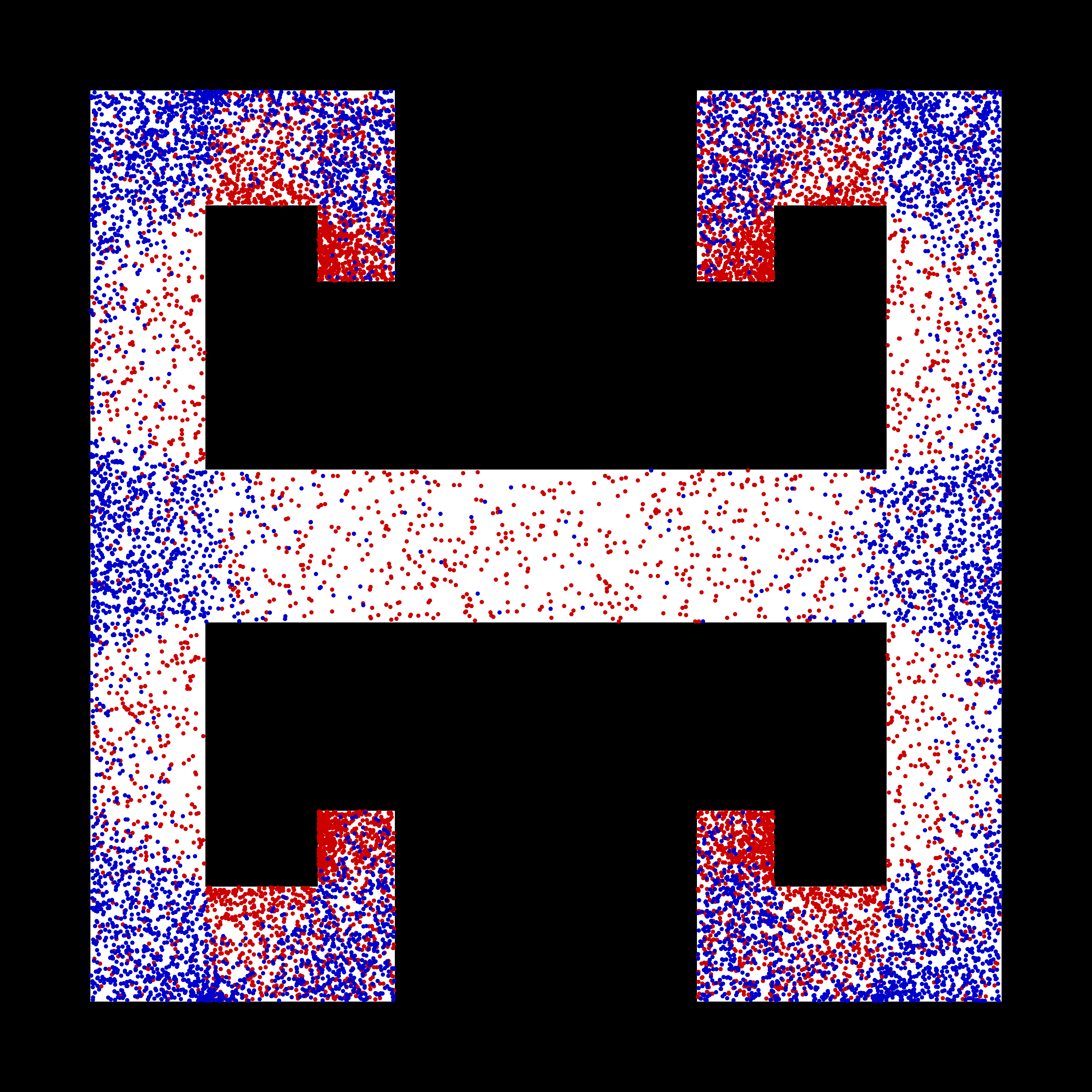}

\caption{[left] An example web.  Red points are initial points; the blue are intersection points.  Green edges connect each intersection point to the initial points from which it is induced.  [right]  The distribution induced by web sampling, based upon 750 generated webs.  Note the higher density at junctions and corners.}
    \label{fig:web-combo}
\end{figure}

The approach is based on a randomized structure called a \emph{web} in $E$, which is constructed in two steps.
First, we construct a set of \emph{initial points} $P \subset E$.  Each initial point is selected sequentially and randomly, from the region outside the union of visibility polygons of the previously selected points.  The process continues until $\bigcup_{p \in P} V(p) = E$.
Then the algorithm selects a collection of \emph{intersection points} $Q \subseteq E$ by examining pairs of initial points $\{p_i, p_j\} \subseteq P$, and placing a point uniformly at random within $V(p_i) \cap V(p_j)$, if that intersection is non-empty.
A web is simply the combination $W = P \cup Q$.
Figure~\ref{fig:web-combo} illustrates this concept.

We utilized webs to generate samples from $E^n$ in Lines~\ref{line:sample1} and \ref{line:sample2} by generating a separate web for each pursuer and sampling, without replacement, from those points. If the points in any of the webs are exhausted, we generate another set of webs and continue.

\gobble{
\begin{algorithm}[t]
  \caption{\textsc{GenerateWebPoints}($E$)}
    \label{algo:web}
  \begin{algorithmic}[1]
    \Input{an environment $E$}
    \State $P \gets \emptyset$
    \State $Q \gets \emptyset$
    \State $F \gets E$
    \While{$F$ contains a shadow}
        \State $p \gets F.\textsc{RandomPoint()}$ 
        \State $P \gets P \cup \{p\}$
        \State $F \gets F \setminus V(p)$
    \EndWhile
    \For{\textbf{each} $p_i, p_j \in P$, with $i<j$}
        \State $A \gets V(p_i) \cap V(p_j) $
        \If{$A \ne \emptyset$}
            \State $q \gets A.\textsc{RandomPoint()}$ 
            \State $Q \gets Q \cup \{q\}$
        \EndIf
    \EndFor
    \State $W \gets P \cup Q$
    \State \Return $W$
  \end{algorithmic}
\end{algorithm}
}

\subsection{Variable numbers of pursuers}\label{ssec:construction}
\label{sec:recursive}\pagebudget{0.75}

In the prior algorithm, the number of pursuers in the solution was required as an input.  
This information was necessary because the SG-PEG data structure stores joint configurations
drawn from $E^n$; thus $n$ must be known to construct an SG-PEG.

To alleviate this limitation, we instead propose a sequential process, in which the number of
pursuers is gradually increased as the algorithm proceeds.
Realizing this approach in the planner requires us to resolve two complications.

First, the algorithm requires a mechanism to transition, in mid-stream, from an $n$-pursuer SG-PEG to
an $(n+1)$-pursuer SG-PEG (Line~\ref{line:clone}).  One straightforward approach is to simply discard
the existing vertices and edges and restart the search with an additional pursuer.  
(This is referred to as the `Clear' option in Section~\ref{sec:experiments}.)

However, it may be preferable to ensure that our previous effort is not wasted.
To this end, we propose a new method that \emph{clones} the first pursuer in each vertex of the SG-PEG.
That is, for each vertex in the SG-PEG, we replace its joint configuration $(p_1, p_2, \dots, p_n)$
with $(p_1, p_2, \dots, p_n, p_1)$. This adds an additional pursuer to the graph, without changing any
of the reachable labels or edges.  Thus, the cloning option is extremely efficient, but it leads to an
SG-PEG in which the newly-added pursuer moves in parallel with another pursuer.  Future edges 
added at the $(n+1)$-th layer will correspond to independent motions for these two robots (as they draw from their own unique set of webs).

Next, we must decide when to expand the number of pursuers (Line~\ref{line:term}). We consider two options.
First, we propose a method that devotes \emph{fixed effort} to each stage of the search.  The process begins by rapidly generating a trivial solution by placing pursuers until their visibility fully covers the environment, which results in a solution that requires no movement from the pursuers.  This gives a (generally very loose) upper bound $N$ on the number of pursuers required.
Then, given a target total run time of $T_{\text{limit}}$ seconds, we apportion
the time between the possible numbers of pursuers $1,\ldots,N$ via a Poisson distribution.  
The Poisson distribution was selected due to the placement of the mean, as well as its skew and shape.
We choose a tunable parameter $\alpha$ which determines the fraction of time to spend on the final step that utilizes $N$ pursuers, so that according to the definition of the Poisson distribution, we have
    $\alpha = \lambda^N e^{-\lambda} / N!$.
From this equation, the algorithm numerically computes the Poisson parameter $\lambda$ and allocates $T_{\text{limit}} \lambda^{i-1} e^{-\lambda}/(i-1)!$ to the search with $i$ pursuers before proceeding to $i+1$.
Because of the existence of the upper bound $N$, this method will always produce a solution, although it may require a large number of pursuers.

As an alternative, we also consider a \textbf{stalled progress} approach, based upon monitoring the
minimum sum of the contaminated shadow area across all vertices of the graph.  (n.b. We have a solution
if and only if this value reaches $0$.)  If this value fails to improve by at least 5\% after adding $M$
samples, we add a new pursuer.  To enable a fair comparison to the fixed effort method, we once again
return a trivial solution if no solution is found with fewer pursuers.

\subsection{Solution refinement}\label{sec:improve}
\pagebudget{0.5}

Because of the sampling-based nature of this algorithm, its outputs are likely to have extraneous motions.
This issue is noticeably more severe in our context than for traditional sampling-based motion planning
because the generated solutions may travel several times along certain edges in an effort to clear
specific shadows.
In this section, we introduce a post processing method which takes a joint motion strategy and optimizes
it by removing unnecessary pursuer motions. 

\begin{figure}[t]
\centering
\resizebox{\columnwidth}{!}{
\begin{tikzpicture}[scale=1, yscale=0.55, xscale=0.35]

\node[fill=black, circle, draw=black, scale=0.5, label=below:{$z_1$}] (z1) at (0,1) {};
\node[fill=black, circle, draw=black, scale=0.5, label=left:{$z_a$}] (za) at (2,2) {};
\node[fill=black, circle, draw=black, scale=0.5] (x) at (4,3.2) {};
\node[fill=black, circle, draw=black, scale=0.5] (y) at (5,1) {};
\node[fill=black, circle, draw=black, scale=0.5] (z) at (7,3.5) {};
\node[fill=black, circle, draw=black, scale=0.5, label=right:{$z_b$}] (zb) at (9,3) {};
\node[fill=black, circle, draw=black, scale=0.5, label=below:{$z_i$}] (zi) at (11,1) {};

\node[fill=black, circle, draw=black, scale=0.5, label=below:{$z_1$}] (z1_2) at (14,1) {};
\node[fill=black, circle, draw=black, scale=0.5, label=left:{$z_a$}] (za_2) at (16,2) {};
\node[fill=black, circle, draw=black, scale=0.5, label=right:{$z_b$}] (zb_2) at (23,3) {};
\node[fill=black, circle, draw=black, scale=0.5, label=below:{$z_i$}] (zi_2) at (25,1) {};

\node[fill=none, label=left:{$c$}] (c) at (7.65,2) {};
\node[fill=none, label=above:{$c'$}] (c') at (19.5,2.5) {};

\draw [color=black,dashed] (za)--(zb);
\draw [color=black] (za)--(x)--(y)--(z)--(zb);
\draw [color=black] (z1)--(za);
\draw [color=black] (zb)--(zi);

\draw [color=cyan] (za_2)--(zb_2);
\draw [color=black] (z1_2)--(za_2);
\draw [color=black] (zb_2)--(zi_2);

\end{tikzpicture}}
\caption{A cut of length $c$ during solution refinement.}
\label{fig:improve}
\end{figure}
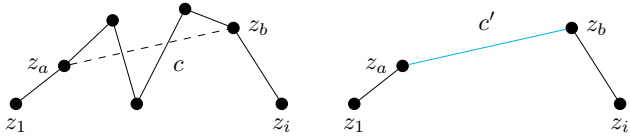

Our method is similar to standard shortcut-based path smoothing.  We select two points $z_a$ and $z_b$ at distance $c$ along the solution path in $E^n$, and check whether taking a shortcut
directly from $z_a$ to $z_b$ yields a path that is still a correct solution.   Figure~\ref{fig:improve}
depicts the process.
Our check features one important difference from traditional path smoothing: In addition to ensuring that
the refined solution is collision-free, we must also ensure that it remains a correct solution, i.e. that
the refinement does not allow any shadows to remain contaminated.  We do this by tracking forward through
the shortened path, applying the shadow events experienced along that path ---Recall
Section~\ref{ssec:shadows}--- to update the shadow labels.

\begin{figure}[t]
    \includegraphics[width=0.47\columnwidth]{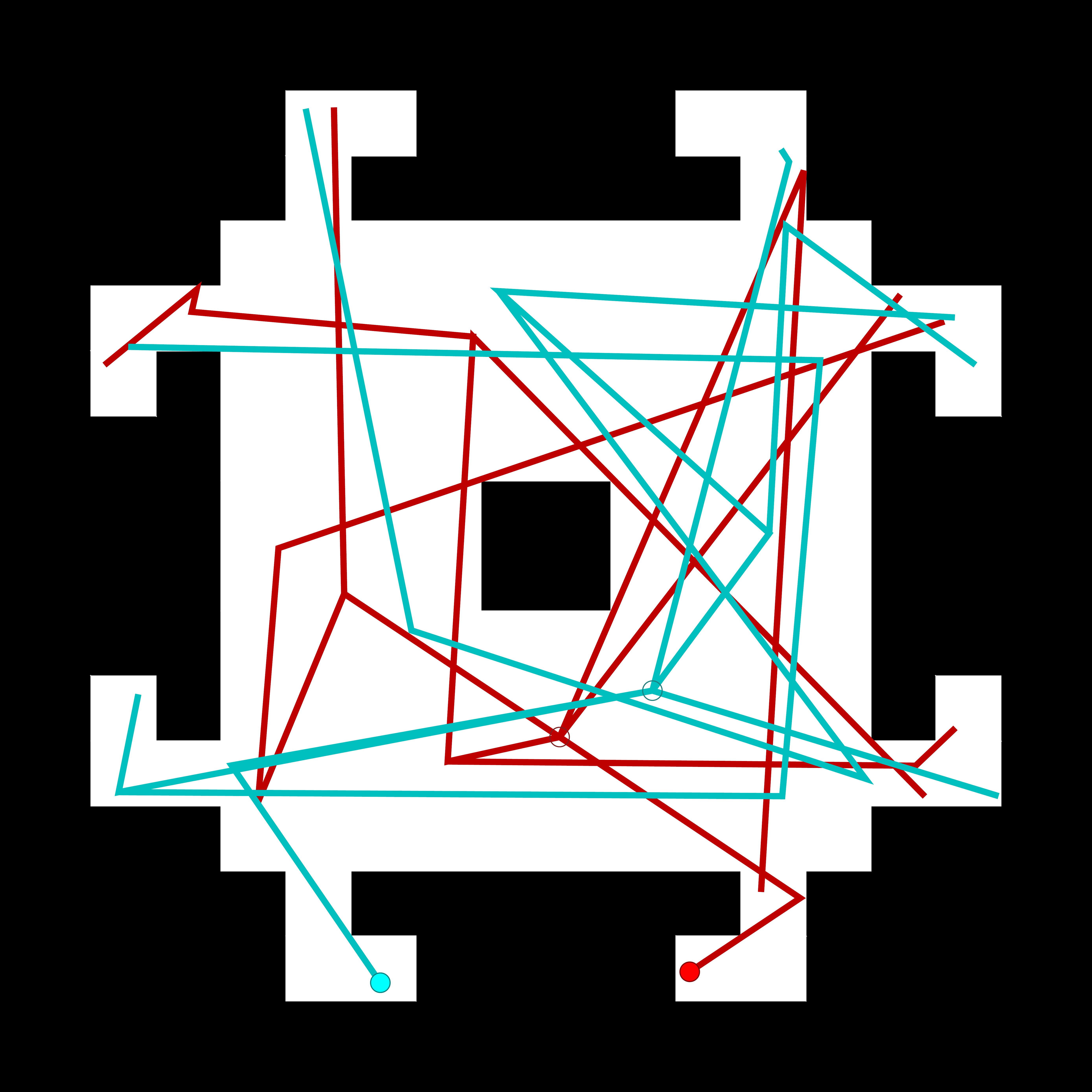}
    \includegraphics[width=0.47\columnwidth]{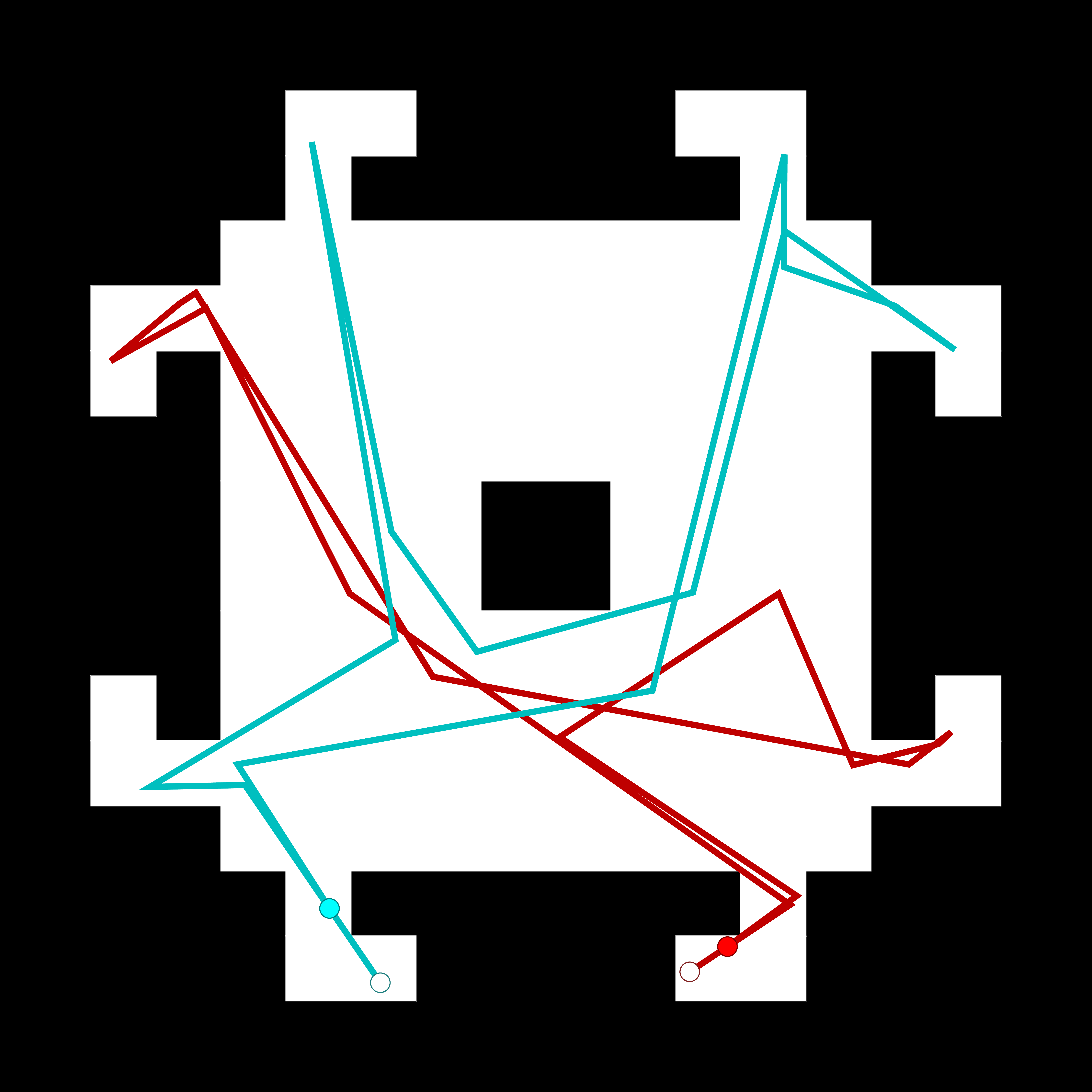}

\caption{Refining a solution.  [left] Before.  [right] After.}
\label{fig:beforeandafter}
\end{figure}

Given this shortcut operation, we greedily optimize the path, proceeding systematically over
decreasing values of $c$ and increasing positions of $z_a$.  (From these two, $z_b$ is readily computed.)
Each time we discover a shortcut yielding a correct solution, that shortened solution replaces the
previous solution, and the process continues.  See Figure~\ref{fig:beforeandafter}.

\gobble{
The nature of sampling will often yield quick, but non-optimized solutions. In the context of the
pursuit-evasion problem,
the solutions generated by the Algorithm in Section~\ref{ssec:construction} may generate a joint motion
strategy 
where some of the pursuers' undertake motions that do not advance the state of the search  
by providing additional information, and thus presents an opportunity to refine the solution. to decrease the travel distance.
Rather than viewing solutions as a collection of pursuer motion strategies in $E \subset \Re^2$, we consider the joint
pursuer strategy as points in $E^n \subset \Re^{2n}$. A solution $X$ can be represented as a path $P$ in $E^n$. Let us
denote the path, of length $k$, starting at $z_1$ by $P=z_1 z_2 \dots z_k$.
Our first improvement comes from finding the smallest index $i$ such that $Q = z_1 \dots z_i, i \leq k$ is a solution . That is,
following the joint pursuer configuration in $E^n$ from $z_1 \dots z_i$ yields a shadow label where all of the shadows are \textit{cleared},
and no further vertices can be removed from the end of the path while still remaining a solution.
We next consider removing unnecessary movements within the solution path $Q=z_1 \cdots z_i$. Let us denote the length
of a cut by $c$. Starting with $c$ being the length of the solution path and reducing $c$ by an arbitrary amount, we
find vertices $z_a$ and $z_b$ along $Q$ that minimize the positive value returned by $d_Q(z_a, z_b) - c$, where $d_{Q}$ denotes
the distance along the curve $Q$. Next, if the path
$Q'= z_1 \cdots z_a z_b \cdots z_i$ is a solution, we set $Q= Q'$ and repeat the process. Once $c$ reaches 0,
we output the most recent $Q$ path as our improved solution. 
}

\section{Evaluation}\label{sec:experiments}\pagebudget{1.5}

\input{tables/table-9roomwithholes}
\input{tables/table-multipleh}
\input{tables/table-8spider}

This section evaluates the performance of the proposed approach.
We implemented the algorithm in C++ and executed it on a computer with an Intel i7-7500U processor
and 12GB of memory, running Ubuntu 18.04.2 64-bit. 

We performed simulations in the environments shown in Figures~\ref{fig:first_page} (\emph{Office}),
\ref{fig:web-combo} (\emph{H}), and \ref{fig:beforeandafter} (\emph{Spider}).
This selection of environments is intended to provide a variety common environmental traits. The H environment contains narrow corridors, the Spider environment possesses numerous hard to reach areas, and the Office environment is somewhat uniformly spread out, with a complex boundary.  %

We generated solution strategies for these environments using both classes of algorithms described thus far: those that utilize a fixed number of pursuers and those that can vary the number of pursuers.
In the former case, we considered both web sampling (WS) and uniform random sampling (SO14)~\cite{StiOKa14b} and varied the fixed number of robots between 2 and 5.
In the latter case, we considered all four combinations of expansion criterion (fixed effort (FE) or stalled progress (SP)) and graph expansion method (clone pursuer (Clone) or clear progress (Clear)).

For each of these scenarios, we executed 25 trials
and recorded the computation time in seconds, number of pursuers used in the returned solution, as well as the total numbers of vertices and edges generated.
Each experiment was allotted 10 minutes of computation time; if no solution was produced in that time, we considered the trial to be a failure. Failed trials are excluded from the statistics, but it should be noted that if they were included, they would contribute a run time of at least 10 minutes. The results, summarized via the 
means $(\mu)$ and standard deviations $(\sigma)$,
appear in Tables~\ref{tab:multipleh}, \ref{tab:nineroomwithholes} and \ref{tab:8spider}.
A number of conclusions may be drawn from the results.

\paragraph{WS improves performance over SO14}
First, in all of the tested environments, our WS method outperformed SO14 by a wide margin.
The smallest improvement in computation time occurred in the Office environment, perhaps due to
the uniform nature of the environment and the complexity of the geometry calculations.

\paragraph{Clone and Clear appear to be complementary}
For the simulations with a variable number of pursuers, we look to compare the performance between the cloning method and the clear method when expanding a graph. When using FE as the expansion criterion, the results are remarkably similar between the two methods. At first glance, one may assume the cloning method should outperform the clearing method due to the larger amount of information the graph contains. This however, is not always the case, because an increased number of vertices means an increased amount of time to add each new sample, due to the propagation of reachable labels across the graph. When SP was the expansion criterion, clearing has a faster computation time, but also a much higher mean and standard deviation for the number of pursuers. It was often the case that SP with clearing would return a trivial solution, as the algorithm cleared a substantial amount of meaningful calculations performed.

\paragraph{Allowing the number of pursuers to vary incurs only a modest computational cost}
Lastly, we compare the fixed number of pursuers using WS, to the variable number of pursuers methods. For each environment, the run times of the variable number of pursuers were within approximately a factor of 2 of that of the fixed number of pursuers. A portion of this additional time comes from the variable number of pursuers constructing a single pursuer graph, in environments which all require at least 2 pursuers to solve.  %

\smallskip

Finally, for each environment, we refined all 25 solutions generated by WS with 2 pursuers. The results are summarized in Table~\ref{tab:improve}, which shows the mean and standard deviation of the computation time (in seconds) along with the length of the solution path (in meters) before and after it was refined.  The results show that this approach effectively and consistently reduces the path lengths. 
\begin{table}[t]
    \caption{Refinement results for the solutions produced by WS ($n=2$).}
    \vspace{-2mm}
    \centering
    \resizebox{0.84\columnwidth}{!}{
    \begin{tabular}{lr@{\hspace{0.75\tabcolsep}}rr@{\hspace{0.75\tabcolsep}}rr@{\hspace{0.75\tabcolsep}}r}
    & \multicolumn{2}{c}{comp} & \multicolumn{2}{c}{length} & \multicolumn{2}{c}{length} \\
    & \multicolumn{2}{c}{time (s)} & \multicolumn{2}{c}{before (m)} & \multicolumn{2}{c}{after (m)} \\
    & \multicolumn{1}{c}{$\mu$} & \multicolumn{1}{c}{$\sigma$} & \multicolumn{1}{c}{$\mu$} & \multicolumn{1}{c}{$\sigma$} & \multicolumn{1}{c}{$\mu$} & \multicolumn{1}{c}{$\sigma$} \\ \hline
    \textbf{Office} & 29.6 & 37.5 & 254.5 & 114.5 & 136.9 & 97.4 \\
    \textbf{H} & 6.4 & 5.6 & 256.5 & 109.4 & 100.0 & 31.7 \\
    \textbf{Spider} & 50.9 & 61.9 & 360.3 & 222.3 & 168.8 & 139.0 \\ \hline
    \end{tabular}
    }
\label{tab:improve}
\end{table}

\section{Conclusion}\label{sec:conclusion}\pagebudget{0.25}
This paper presents a sampling-based algorithm for a visibility-based pursuit-evasion problem that 
generates a joint motion strategy for a team of robots in a polygonal environment. 
The three primary contributions are a novel sampling strategy for this domain,
an iterative algorithm for generating a joint motion strategy for the pursuers, and a post-processing
path-smoothing algorithm that refines the strategy returned by the main algorithm.
The algorithm was shown to outperform existing techniques.

Future work might build upon the results in this paper.
First, possibilities remain for enhancing the post processing step.  
There remain a number of open questions on how these kinds of path smoothing algorithms can be best applied to the pursuit-evasion domain.
Second, is the development of an anytime algorithm that begins with an uninteresting solution 
---for example, one utilizing enough pursuers to ensure that their visibility polygons fully
cover the environment--- and attempts to work backwards, eliminating robots, by searching for solutions
in the reduced joint sub-space.
\gobble{For example, if the objective is to reduce the total amount of travel by all of the pursuers (rather than, as considered here, the travel required by the longest pursuer path), then local modifications in which the motions are shortened for only a subset of the pursuers may prove effective.}
\gobble{Another possible thread of future work is the development of an anytime algorithms that begin
with an uninteresting solution ---for example, one utilizing enough pursuers to ensure that their
visibility polygons fully cover the environment---, and incrementally attempts to remove pursuers, adding
additional vertices to explore the space more fully. Such algorithms would certainly be able to emit
correct solutions rapidly, and may also be able to generate a solution using few robots, given an adequate
computation time.}

\pagebreak

\cleardoublepage

\begin{refcontext}[sorting=nyt]
\printbibliography
\end{refcontext}

\end{document}

%% file: tables/table-9roomwithholes.tex
\begin{table*}[t]
    \caption{Simulation results for the Office environment (Figure~\ref{fig:first_page}).}
    \vspace{-2mm}
    \centering
    \resizebox{0.89\textwidth}{!}{
    \begin{tabular}{lcr@{\hspace{0.75\tabcolsep}}rr@{\hspace{0.75\tabcolsep}}rr@{\hspace{0.75\tabcolsep}}rr@{\hspace{0.75\tabcolsep}}r}
    & success & \multicolumn{2}{c}{comp time (s)} & \multicolumn{2}{c}{num robots} & \multicolumn{2}{c}{num vertices} & \multicolumn{2}{c}{num edges} \\
    & \multicolumn{1}{c}{rate} & \multicolumn{1}{c}{$\mu$} & \multicolumn{1}{c}{$\sigma$} & \multicolumn{1}{c}{$\mu$} & \multicolumn{1}{c}{$\sigma$} & \multicolumn{1}{c}{$\mu$} & \multicolumn{1}{c}{$\sigma$} & \multicolumn{1}{c}{$\mu$} & \multicolumn{1}{c}{$\sigma$} \\ \hline
    \textbf{WS \ \ ($n=2$)} & 100\% & 87.201 & 58.707 & 2.000 & 0.000 & 74.360 & 50.392 & 137.440 & 103.644 \\
    \textbf{SO14 ($n=2$)} & 96\% & 96.890 & 78.231 & 2.000 & 0.000 & 117.875 & 62.007 & 219.167 & 131.542  \\ \hline
    \textbf{WS \ \ ($n=3$)} & 100\% & 68.953 & 56.738 & 3.000 & 0.000 & 109.160 & 55.984 & 157.040 & 105.265 \\
    \textbf{SO14 ($n=3$)} & 100\% & 63.205 & 25.110 & 3.000 & 0.000 & 330.440 & 210.853 & 542.080 & 453.409  \\ \hline
    \textbf{WS \ \ ($n=4$)} & 100\% & 57.248 & 52.329 & 4.000 & 0.000 & 177.760 & 73.723 & 167.960 & 103.567 \\
    \textbf{SO14 ($n=4$)} & 100\% & 73.448 & 65.325 & 4.000 & 0.000 & 1258.480 & 774.096 & 1830.920 & 1655.530  \\ \hline
    \textbf{WS \ \ ($n=5$)} & 100\% & 91.633 & 80.529 & 5.000 & 0.000 & 392.520 & 229.407 & 275.920 & 224.627 \\
    \textbf{SO14 ($n=5$)} & 80\% & 235.866 & 129.786 & 5.000 & 0.000 & 4081.000 & 1361.404 & 4495.150 & 2546.067  \\ \hline
    \textbf{FE Clone $(\alpha = 0.001)$} & 100\% & 138.097 & 71.992 & 3.440 & 0.768 & 138.720 & 88.246 & 186.720 & 89.464 \\
    \textbf{FE Clear $(\alpha = 0.001)$} & 100\% & 111.114 & 112.003 & 3.280 & 2.390 & 866.360 & 2702.653 & 732.200 & 1379.524  \\
    \textbf{SP Clone $(M=30)$} & 100\% & 193.867 & 55.951 & 2.520 & 2.365 & 117.280 & 79.072 & 192.200 & 120.596 \\
    \textbf{SP Clear $(M=30)$} & 100\% & 84.824 & 29.878 & 7.640 & 4.636 & 222.360 & 107.938 & 185.160 & 35.873  \\ \hline

    \end{tabular}
      }
\label{tab:nineroomwithholes}
\end{table*}

%% file: tables/table-multipleh.tex
\begin{table*}[t]
    \caption{Simulation results for the H environment (Figure~\ref{fig:web-combo}).}
    \vspace{-2mm}
    \centering
    \resizebox{0.89\textwidth}{!}{
    \begin{tabular}{lcr@{\hspace{0.75\tabcolsep}}rr@{\hspace{0.75\tabcolsep}}rr@{\hspace{0.75\tabcolsep}}rr@{\hspace{0.75\tabcolsep}}r}
    & success & \multicolumn{2}{c}{comp time (s)} & \multicolumn{2}{c}{num robots} & \multicolumn{2}{c}{num vertices} & \multicolumn{2}{c}{num edges} \\
    & \multicolumn{1}{c}{rate} & \multicolumn{1}{c}{$\mu$} & \multicolumn{1}{c}{$\sigma$} & \multicolumn{1}{c}{$\mu$} & \multicolumn{1}{c}{$\sigma$} & \multicolumn{1}{c}{$\mu$} & \multicolumn{1}{c}{$\sigma$} & \multicolumn{1}{c}{$\mu$} & \multicolumn{1}{c}{$\sigma$} \\ \hline
    \textbf{WS \ \ ($n=2$)} & 100\% & 18.716 & 9.816 & 2.000 & 0.000 & 71.600 & 49.019 & 119.160 & 103.625 \\
    \textbf{SO14 ($n=2$)} & 100\% & 33.885 & 14.777 & 2.000 & 0.000 & 105.760 & 77.638 & 193.640 & 158.844  \\ \hline
    \textbf{WS \ \ ($n=3$)} & 100\% & 20.151 & 11.281 & 3.000 & 0.000 & 143.200 & 46.871 & 168.720 & 84.149 \\
    \textbf{SO14 ($n=3$)} & 100\% & 35.965 & 20.354 & 3.000 & 0.000 & 212.560 & 152.480 & 349.840 & 305.536  \\ \hline
    \textbf{WS \ \ ($n=4$)} & 100\% & 27.131 & 10.502 & 4.000 & 0.000 & 261.520 & 94.841 & 211.920 & 116.173 \\
    \textbf{SO14 ($n=4$)} & 96\% & 43.636 & 42.848 & 4.000 & 0.000 & 639.042 & 365.724 & 972.042 & 704.745  \\ \hline
    \textbf{WS \ \ ($n=5$)} & 100\% & 47.967 & 24.342 & 5.000 & 0.000 & 413.840 & 119.927 & 196.800 & 77.066 \\
    \textbf{SO14 ($n=5$)} & 96\% & 81.949 & 62.208 & 5.000 & 0.000 & 1743.708 & 1512.375 & 2422.125 & 2732.005  \\ \hline
    \textbf{FE Clone $(\alpha = 0.001)$} & 100\% & 27.065 & 9.973 & 2.000 & 0.000 & 48.120 & 16.269 & 79.440 & 31.466 \\
    \textbf{FE Clear $(\alpha = 0.001)$} & 100\% & 21.538 & 8.483 & 2.000 & 0.000 & 71.360 & 24.124 & 115.520 & 48.321  \\
    \textbf{SP Clone $(M=30)$} & 100\% & 42.422 & 13.620 & 2.400 & 1.607 & 76.120 & 29.992 & 129.320 & 46.703 \\
    \textbf{SP Clear $(M=30)$} & 100\% & 19.637 & 6.906 & 7.920 & 3.081 & 204.840 & 62.487 & 136.640 & 31.579  \\ \hline

    \end{tabular}
    }
\label{tab:multipleh}
\end{table*}

%% file: tables/table-8spider.tex
\begin{table*}[t]
    \caption{Simulation results for the Spider environment (Figure~\ref{fig:beforeandafter}).}
    \vspace{-2mm}
    \centering
    \resizebox{0.89\textwidth}{!}{
    \begin{tabular}{lcr@{\hspace{0.75\tabcolsep}}rr@{\hspace{0.75\tabcolsep}}rr@{\hspace{0.75\tabcolsep}}rr@{\hspace{0.75\tabcolsep}}r}
    & success & \multicolumn{2}{c}{comp time (s)} & \multicolumn{2}{c}{num robots} & \multicolumn{2}{c}{num vertices} & \multicolumn{2}{c}{num edges} \\
    & \multicolumn{1}{c}{rate} & \multicolumn{1}{c}{$\mu$} & \multicolumn{1}{c}{$\sigma$} & \multicolumn{1}{c}{$\mu$} & \multicolumn{1}{c}{$\sigma$} & \multicolumn{1}{c}{$\mu$} & \multicolumn{1}{c}{$\sigma$} & \multicolumn{1}{c}{$\mu$} & \multicolumn{1}{c}{$\sigma$} \\ \hline
    \textbf{WS \ \ ($n=2$)} & 100\% & 151.738 & 87.515 & 2.000 & 0.000 & 54.240 & 34.170 & 90.920 & 65.193 \\
    \textbf{SO14 ($n=2$)} & 84\% & 312.341 & 115.955 & 2.000 & 0.000 & 152.952 & 90.760 & 296.619 & 182.206  \\ \hline
    \textbf{WS \ \ ($n=3$)} & 100\% & 91.825 & 41.326 & 3.000 & 0.000 & 50.120 & 36.003 & 68.280 & 58.533 \\
    \textbf{SO14 ($n=3$)} & 96\% & 211.234 & 89.930 & 3.000 & 0.000 & 90.167 & 33.983 & 160.750 & 65.811  \\ \hline
    \textbf{WS \ \ ($n=4$)} & 100\% & 84.356 & 26.550 & 4.000 & 0.000 & 55.520 & 26.590 & 61.880 & 40.169 \\
    \textbf{SO14 ($n=4$)} & 96\% & 216.941 & 83.815 & 4.000 & 0.000 & 142.708 & 283.093 & 257.083 & 560.368  \\ \hline
    \textbf{WS \ \ ($n=5$)} & 100\% & 92.822 & 42.587 & 5.000 & 0.000 & 75.800 & 33.582 & 72.920 & 47.970 \\
    \textbf{SO14 ($n=5$)} & 92\% & 222.129 & 77.185 & 5.000 & 0.000 & 81.261 & 31.517 & 123.087 & 55.083  \\ \hline
    \textbf{FE Clone $(\alpha = 0.001)$} & 100\% & 249.742 & 116.423 & 5.080 & 1.631 & 107.720 & 40.658 & 117.200 & 46.227 \\
    \textbf{FE Clear $(\alpha = 0.001)$} & 100\% & 258.762 & 124.280 & 5.080 & 2.159 & 196.200 & 188.685 & 184.200 & 110.659  \\
    \textbf{SP Clone $(M=30)$} & 88\% & 403.098 & 91.530 & 3.727 & 0.827 & 163.682 & 35.301 & 275.955 & 50.614 \\
    \textbf{SP Clear $(M=30)$} & 100\% & 236.300 & 134.572 & 4.200 & 4.021 & 139.680 & 102.347 & 184.440 & 65.614  \\ \hline

    \end{tabular}
    }
\label{tab:8spider}
\end{table*}